\documentclass[sigconf]{acmart}

\usepackage{amsmath}    
\usepackage{algorithm}
\usepackage{algorithmic}
\usepackage{enumitem}
\usepackage{multirow}
\usepackage{booktabs}   
\usepackage{array}      
\usepackage{subcaption} 
\usepackage{xcolor}

\AtBeginDocument{%
  }

\setcopyright{acmlicensed}
\copyrightyear{2018}
\acmYear{2018}
\acmDOI{XXXXXXX.XXXXXXX}
\acmConference[Conference acronym 'XX]{Make sure to enter the correct
  conference title from your rights confirmation email}{June 03--05,
  2018}{Woodstock, NY}
\acmISBN{978-1-4503-XXXX-X/2018/06}

\begin{document}

\title{HeteroBA: A Structure-Manipulating Backdoor Attack on Heterogeneous Graphs}


\author{Honglin Gao}
\affiliation{%
  \department{School of Electrical and Electronic Engineering}
  \institution{Nanyang Technological University}
  \country{Singapore}}
\email{honglin001@e.ntu.edu.sg}

\author{Xiang Li}
\affiliation{%
  \department{School of Electrical and Electronic Engineering}
  \institution{Nanyang Technological University}
  \country{Singapore}}
\email{xiang002@e.ntu.edu.sg}

\author{Lan Zhao}
\affiliation{%
  \department{School of Electrical and Electronic Engineering}
  \institution{Nanyang Technological University}
  \country{Singapore}}
\email{zhao0468@e.ntu.edu.sg}

\author{Gaoxi Xiao}
\affiliation{%
  \department{School of Electrical and Electronic Engineering}
  \institution{Nanyang Technological University}
  \country{Singapore}}
\email{egxxiao@ntu.edu.sg}






\renewcommand{\shortauthors}{Trovato et al.}

\begin{abstract}
  Heterogeneous graph neural networks (HGNNs) have recently drawn increasing attention for modeling complex multi-relational data in domains such as recommendation, finance, and social networks. While existing research has been largely focusing on enhancing HGNNs’ predictive performance, their robustness and security, especially under backdoor attacks, remain underexplored. In this paper, we propose a novel Heterogeneous Backdoor Attack (HeteroBA) framework for node classification tasks on heterogeneous graphs. HeteroBA inserts carefully crafted trigger nodes with realistic features and targeted structural connections, leveraging attention-based and clustering-based strategies to select influential auxiliary nodes for effective trigger propagation, thereby causing the model to misclassify specific nodes into a target label while maintaining accuracy on clean data. Experimental results on three datasets and various HGNN architectures demonstrate that HeteroBA achieves high attack success rates with minimal impact on the clean accuracy. Our method sheds light on potential vulnerabilities in HGNNs and calls for more robust defenses against backdoor threats in multi-relational graph scenarios.
\end{abstract}


\received{20 February 2007}
\received[revised]{12 March 2009}
\received[accepted]{5 June 2009}

\maketitle

\section{Introduction}
Graph data is prevalent in various applications, including social networks \citep{gnnsocialnet,GNNSocialNet2}, signal processing \citep{gnnsingalprocess}, biological networks \citep{gnnbiological}, and knowledge graphs \citep{gnnknowledgegraph}. Unlike homogeneous graphs, heterogeneous graphs (HGs) contain multiple node and edge types, making them particularly effective for modeling complex real-world relationships. For instance, an academic graph comprises researchers, papers, and institutions connected through authorship, citation, and collaboration links. Such structural flexibility allows heterogeneous graphs to serve as the backbone for various domains, including recommendation systems and financial risk modeling \citep{recommandsystem, financeriskassesment}, etc.

Heterogeneous Graph Neural Networks (HGNNs) extend GNNs to incorporate diverse relational information, making them well-suited for tasks like node classification \citep{HGNNNC1, HGNNNC2} and link prediction \citep{HGNNLP1, HGNNLP2}. In financial applications, HGNNs have been utilized for fraud detection and risk assessment \citep{heteorgeneousFinance1, heteorgeneousFinance2}, while in recommender systems, they enhance personalized recommendations by capturing cross-domain interactions \citep{heterogeneousRecommendationSystem1, heterogeneousRecommendationSystem2}. Despite these advantages, research has primarily focused on improving HGNN performance, leaving their security vulnerabilities relatively underexplored. Recent studies highlight their susceptibility to adversarial threats \citep{heterogeneousGraphAttack1, heterogeneousGraphAttack2}, among which backdoor attacks pose a particularly severe risk due to their stealthiness and potential impact on critical decision-making \citep{GraphBackdoor}.

Backdoor attacks aim to manipulate a model’s behavior under specific input conditions by intentionally altering its training data. While traditional backdoor attacks have been extensively studied in domains such as computer vision \cite{CVBackdoor1, CVBackdoor2} and natural language processing \cite{nlpBackdoor1,nlpBackdoor2}, research on backdoor attacks in HGNNs remains scarce. Unlike homogeneous graphs, heterogeneous graphs capture richer structural and semantic information through diverse node types and relationships. HGNNs leverage these heterogeneous connections to learn comprehensive feature representations for downstream tasks. However, the complexity and diversity of heterogeneous graphs also introduce new vulnerabilities, providing attackers with opportunities to embed backdoors by carefully modifying the graph structure or feature representations in a targeted manner. Once a backdoor is successfully implanted, an attacker can exploit specific trigger patterns to mislead the model into producing incorrect outputs, potentially leading to severe security risks. For example, in financial systems, an attacker can introduce a fake bank account under a customer's identity into the network. By creating hidden connections between the fraudulent account and transaction history, the attacker can manipulate the system’s judgment, potentially evading detection. 

 Existing backdoor attack methods primarily focus on homogeneous graphs. Dai et al. argue that trigger insertion disrupts the message-passing process between nodes, thereby compromising the model’s integrity \cite{UGBA}. To address this, they propose Unnoticeable Graph Backdoor Attack (UGBA), which leverages a bi-level optimization framework to execute backdoor attacks under a limited attack budget while minimizing detectability. In contrast, Xing et al. introduce Clean-Label Graph Backdoor Attack (CGBA) \cite{CGBA}, which injects triggers into node feature representations without altering node labels or graph structure. By selecting triggers from existing node features with high similarity to their neighbors, CGBA enhances attack stealthiness and avoids structural modifications, making it more resistant to defense mechanisms. However, these methods of backdoor attacks on node classification tasks overlook the diversity of edge relationships and node types. To address this limitation and exploit potential vulnerabilities in HGNNs, we propose a novel \underline{Hetero}geneous \underline{B}ackdoor \underline{A}ttack (HeteroBA) method specifically designed for node classification in heterogeneous graphs.

Unlike existing node classification backdoor attack methods, HeteroBA represents a novel and targeted approach specifically engineered for heterogeneous graphs. By strategically injecting minimally invasive triggers, HeteroBA effectively achieves superior attack performance while maintaining exceptional invisibility, setting a new standard in graph-based adversarial techniques.

Specifically, when the targeted nodes of the attack have been selected, new trigger nodes are introduced into the graph to carry out the attack. These trigger nodes are strategically connected to the targeted nodes and some highly influential nodes, i.e., those nodes that can significantly impact information propagation within the graph, forming subtle but effective perturbations. The features of the trigger nodes are constructed based on the statistical properties of nodes of the same type, ensuring consistency with the existing graph structure and enhancing stealthiness. Finally, after modifying the labels of the targeted nodes, the poisoned graph, embedded with these adversarial triggers and connections, ensures that when the backdoored model encounters similar trigger patterns during inference, it misclassifies the targeted nodes as belonging to a specific wrongful class as intended by the attacker. Note that the proposed attack can lead to misclassification of multiple targeted nodes into the same wrongful class (hereafter termed as {\it designated target class}). 

The main contribution of this paper is as follows:
\begin{itemize}
    \item We propose HeteroBA, the first dedicated backdoor attack on heterogeneous graphs for the node classification task, achieving high attack success rates across various models and datasets.
    \item HeteroBA effectively manipulates the graph structure to enhance the attack while maintaining efficiency in execution.
    \item Extensive experiments on multiple benchmark datasets validate the effectiveness of HeteroBA, outperforming baselines in multiple cases and demonstrating strong attack capability. We also propose a method to calculate the stealthiness score for node injection-based backdoor attacks.
\end{itemize}

The remainder of this paper is structured as follows: Section 2 provides a brief review of related work. Section 3 introduces the necessary preliminaries and definitions. Our proposed methodology is presented in detail in Section 4. To evaluate its effectiveness, we conduct extensive experiments and analyses on multiple benchmark datasets and models in Section 5. Finally, Section 6 concludes the paper. Our code has been open-sourced and is publicly available \footnote{https://anonymous.4open.science/r/HeteroBA-EEAF}.

\section{Related Work}

\subsection{Heterogeneous Graph Neural Networks}
HGNNs have evolved significantly in recent years \citep{RHGNN,HGCN,MAGNN,RpHGNN}, with various architectures designed to effectively capture heterogeneous relationships. Among them, the most representative models include a few as follows. HAN \citep{HAN} introduces meta-path-based attention to selectively aggregate information along predefined relational paths, providing interpretability in node representations. HGT \citep{HGT} extends this approach by leveraging a transformer-based architecture to dynamically model heterogeneous interactions. Meanwhile, SimpleHGN \citep{SimpleHGN} optimizes message passing by simplifying the heterogeneity modeling process, making it computationally efficient while maintaining strong performance. 

While these methods significantly improve learning on heterogeneous graphs, their robustness under malicious manipulation has received limited attention. The unique characteristics of HGNNs, such as diverse node and edge types and advanced attention mechanisms, present both opportunities and challenges for potential attackers, making them an important area for further exploration.

\subsection{Backdoor Attack on Homogeneous Graph}
Backdoor attacks on graph neural networks (GNNs) embed hidden triggers during training, allowing adversaries to control outputs under specific conditions. Existing attacks on homogeneous graphs are classified by their trigger injection strategies.

\textbf{Feature-based Backdoor Attacks} introduce malicious triggers by modifying node attributes while keeping the graph structure unchanged. NFTA (Node Feature Target Attack) \citep{NFTA} injects feature triggers without requiring knowledge of GNN parameters, disrupting the feature space and confusing model predictions. It also introduces an adaptive strategy to balance feature smoothness. Xing et al. \citep{CGBA} selected trigger nodes with high similarity to neighbors, ensuring stealthiness without modifying labels or structure. However, both methods rely solely on feature manipulation, making them less effective when structural changes significantly impact message passing.

\textbf{Structure-based Backdoor Attacks} manipulate the graph topology by adding or removing edges to implant triggers. Zhang et al. \citep{zhang2021backdoor} introduced a subgraph-based trigger to mislead graph classification models while maintaining high attack success rates. Xi et al. \citep{GTA} extended this concept by generating adaptive subgraph triggers that dynamically tailor the attack for different inputs. Dai et al. \citep{UGBA} employed a bi-level optimization strategy to modify graph structures under an attack budget, maximizing stealthiness while ensuring effectiveness.

Although these methods demonstrate the feasibility of backdoor attacks on homogeneous graphs, their reliance on uniform graph structures and simple node relationships limits their applicability to the more complex heterogeneous graph setting.

\subsection{Attacks on Heterogeneous Graph Neural Networks}
Heterogeneous graphs have shown vulnerabilities under adversarial attacks, and several studies have explored this area. Zhang et al. \citep{RoHe} proposed RoHe, a robust HGNN framework that defends against adversarial attacks by pruning malicious neighbors using an attention purifier. Zhao et al. \citep{HGAttack} introduced HGAttack, the first grey-box evasion attack specifically targeting HGNNs, which leverages a semantic-aware mechanism and a novel surrogate model to generate perturbations. These works highlight the susceptibility of HGNNs to adversarial manipulations and the progress made in addressing these threats.

However, while adversarial attacks on heterogeneous graphs and backdoor attacks on homogeneous graphs have been explored, no prior work has investigated backdoor vulnerabilities in HGNNs. Our research addresses this gap by proposing a novel backdoor attack method specifically designed for heterogeneous graphs, leveraging their unique structural properties to embed triggers while maintaining high attack success rates and stealthiness.

\section{Preliminaries and problem formulation}
In this section, we introduce the preliminaries of backdoor attacks on heterogeneous graphs and define the problem. Table \ref{tab:notation} summarizes the notation used throughout this section for clarity.

\subsection{Preliminaries}
\textbf{Definition 3.1 (Heterogeneous graph)}
A heterogeneous graph is defined as $G=\{\mathcal{V}, \mathcal{E}, X\}$, where $\mathcal{V}=\left\{v_1, v_2, \ldots, v_n\right\}$ is the node set, and $X \in \mathbb{R}^{|\mathcal{V}| \times d}$ is a node feature matrix with $d$ being the dimension of each node feature.

The set $\mathcal{T}=\left\{t_1, t_2, \ldots, t_T\right\}$ represents $T$ different node types, where each node $v \in \mathcal{V}$ belongs to one specific type $t \in \mathcal{T}$. Nodes for each type $t$ is represented by the subset $\mathcal{V}_t$, and its size is denoted as $\left\|\mathcal{V}_t\right\|$.
The set of edge types is denoted as
$
\mathcal{R}=\left\{r_{t_a, t_b} \mid t_a, t_b \in \mathcal{T}, t_a \neq t_b\right\}
$
where each edge type $r_{t_a, t_b}$ represents connections between nodes of type $t_a$ and nodes of type $t_b$. For each pair of node types $\left(t_a, t_b\right)$, we maintain an adjacency matrix $A_{t_a, t_b} \in\{0,1\}^{\left|\mathcal{V}_{t_a}\right| \times\left|\mathcal{V}_{t_b}\right|}$, where $A_{t_a, t_b}\left(v_i, v_j\right)=1$ indicates an edge between node $v_i \in \mathcal{V}_{t_a}$ and node $v_j \in \mathcal{V}_{t_b}$. We then define the edge set $\mathcal{E}$ as the union of all such edges, recorded as triples $\left(v_i, v_j, r_{t_a, t_b}\right)$ :
$
\mathcal{E}=\bigcup_{r_{t_a, t_b} \in \mathcal{R}}\left\{\left(v_i, v_j, r_{t_a, t_b}\right) \mid v_i \in \mathcal{V}_{t_a}, v_j \in \mathcal{V}_{t_b}, A_{t_a, t_b}\left(v_i, v_j\right)=1\right\}
$
Hence, each adjacency matrix $A_{t_a, t_b}$ describes the connectivity between nodes of types $t_a$ and $t_b$, and each nonzero entry in $A_{t_a, t_b}$ corresponds to an edge in $\mathcal{E}$. A heterogeneous graph satisfies the condition $T+|\mathcal{R}|>2$.

\textbf{Definition 3.2 (Primary type, trigger type and auxiliary type)}
We define three key types. The primary type $t_p \in \mathcal{T}$ refers to 
the type of nodes for classification. The trigger type 
$t_{tr} \in \mathcal{T}$ denotes the type of nodes added as backdoor triggers. 
The auxiliary type comprises node types $t_a \in \mathcal{T}$ that can be 
reached from a primary-type node $v_{t_p} \in \mathcal{V}_{t_p}$ via a 
trigger-type node $v_{t_{tr}} \in \mathcal{V}_{t_{tr}}$ in exactly two hops. 
Formally, $\mathcal{T}_{\mathrm{aux}} 
= \bigl\{ 
   t_a \in \mathcal{T}
   \,\big|\,
   \exists\, v_{t_p} \in \mathcal{V}_{t_p},\, 
            v_{t_{tr}} \in \mathcal{V}_{t_{tr}},\, 
            v_{t_a} \in \mathcal{V}_{t_a} : 
   (v_{t_p}, v_{t_{tr}}, r_{t_p, t_{tr}}) \in \mathcal{E},\,
   (v_{t_{tr}}, v_{t_a}, r_{t_{tr}, t_a}) \in \mathcal{E}
\bigr\}.$

\textbf{Definition 3.3 (Designated target class and non-target classes)}
Let $\mathcal{Y}$ denote the set of class labels in the classification task. The designated target class is defined as $y_t \in \mathcal{Y}$, representing the label to which the attacker aims to misclassify certain nodes. The non-target classes are given by $\mathcal{Y}_{\neg t}=\mathcal{Y} \backslash\left\{y_t\right\}$.
In a heterogeneous graph, classification is performed on nodes of the primary type $t_p$, whose node set is denoted as $\mathcal{V}_{t_p}$. Based on their ground-truth labels, we define:
$
\mathcal{V}_{y_t}=\left\{v \in \mathcal{V}_{t_p} \mid y_v=y_t\right\}, \quad \mathcal{V}_{\neg y_t}=\left\{v \in \mathcal{V}_{t_p} \mid y_v \neq y_t\right\}
$. By definition, $\mathcal{V}_{y_t} \cup \mathcal{V}_{\neg y_t}=\mathcal{V}_{t_p}$ and $\mathcal{V}_{y_t} \cap \mathcal{V}_{\neg y_t}=\emptyset$.

\begin{figure}[t]
    \centering
    \includegraphics[width=1.0\linewidth]{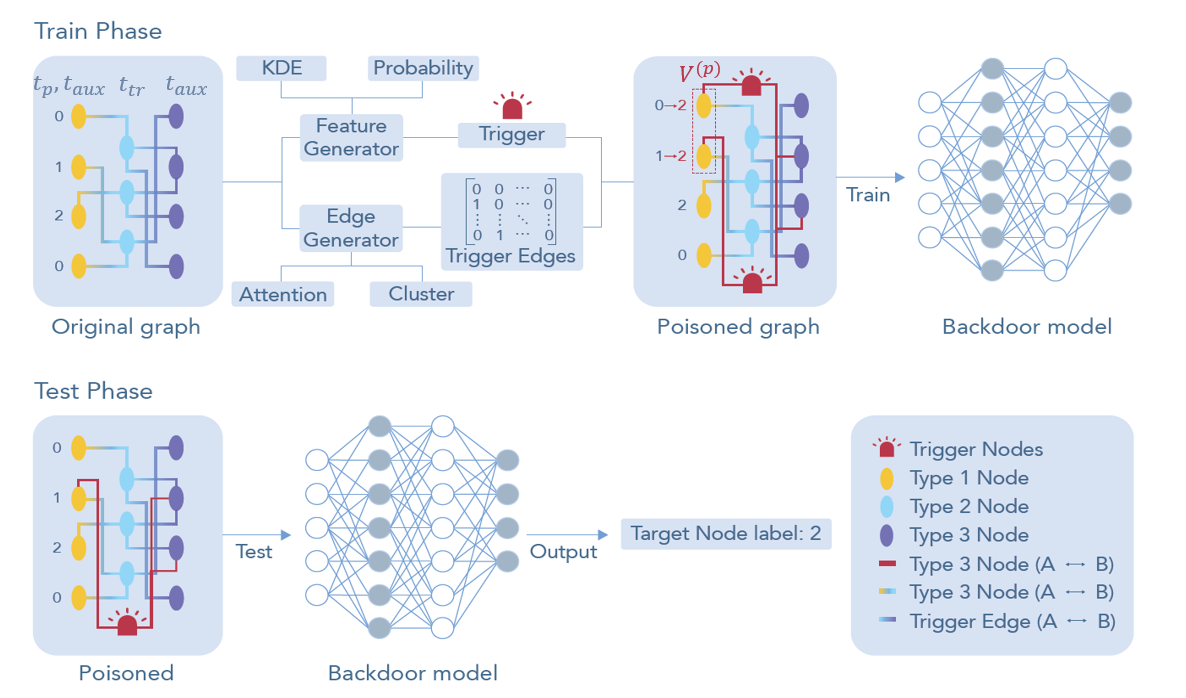}
    \caption{Overall Backdoor Attack Process on a Heterogeneous Graph.}
    \label{fig:wholeprocess}
\end{figure}

\subsection{Problem definition}
Given a heterogeneous graph $G=(\mathcal{V}, \mathcal{E}, X)$ and a node classification model $f_\theta: \mathcal{V}_{t_p} \rightarrow \mathcal{Y}$ trained on the primary node type $\mathcal{V}_{t_p} \subseteq \mathcal{V}$, a backdoor attack alters $G$ to construct a poisoned graph $\widetilde{G}=(\widetilde{\mathcal{V}}, \widetilde{\mathcal{E}}, \widetilde{X})$, ensuring that after training on $\widetilde{G}$, the model misclassifies specific target nodes while preserving overall classification accuracy. To construct $\widetilde{G}$, the attacker introduces a set of new trigger nodes $\mathcal{V}_{t_{t_r}}^{(\text {new})}$ with feature matrix $X^{(\text {new})}$ and new edges $\mathcal{E}^{(n e w)}$ connecting them to existing nodes, resulting in $\widetilde{\mathcal{V}}=\mathcal{V} \cup \mathcal{V}_{t_{\text {tr }}}^{(\text {new })}, \widetilde{\mathcal{E}}=\mathcal{E} \cup \mathcal{E}^{(\text {new })}$, and $\widetilde{X}=\left[\begin{array}{c}X \\ X^{(\text {new })}\end{array}\right]$. Specifically, the attacker selects a subset of primary-type nodes, denoted as $\mathcal{V}^{(p)} \subseteq \mathcal{V}_{t_p}$, as targeted nodes, and aims to enforce their wrongful classification into a designated target class $y_t$ during inference, i.e., $f_\theta(\widetilde{G}, v)=y_t, \forall v \in \mathcal{V}^{(p)}$. Meanwhile, for the remaining primary-type nodes $\mathcal{V}_{t_p} \backslash \mathcal{V}^{(p)}$, the model should retain its correct predictions, i.e., $f_\theta(\widetilde{G}, v)=y_v, \forall v \in \mathcal{V}_{t_p} \backslash \mathcal{V}^{(p)}$. Formally, the attack is formulated as an optimization problem:
\begin{equation}
\scriptsize 
\widetilde{G}^* =
\arg\max_{\widetilde{G} \,\in\, \mathcal{F}(G)}
\left[
    \sum_{v \in \mathcal{V}^{(p)}}
    \mathbf{1}\bigl(f_{\theta}(\widetilde{G}, v) = y_t\bigr)
    +
    \sum_{v \in \mathcal{V}_{t_p} \setminus \mathcal{V}^{(p)}}
    \mathbf{1}\bigl(f_{\theta}(\widetilde{G}, v) = y_v\bigr)
\right]
\end{equation}

Here, $\mathbf{1}(\cdot)$ is an indicator function that returns 1 if the condition is satisfied and 0 otherwise, and $\mathcal{F}(G)$ denotes the space of permissible modifications to $G$, which may include adding or modifying nodes, edges, or node features.

\begin{table}[t]
\small
\caption{Notation and Definitions}
\label{tab:notation}
\centering
\begin{tabular}{cl}
\toprule
Symbol & Meaning \\
\midrule
$G=(\mathcal{V},\mathcal{E},X)$ 
& Heterogeneous graph \\
$\mathcal{V}, \mathcal{E}, X$ 
& Nodes, edges, feature matrix \\
$\mathcal{T}, \mathcal{R}$ 
& Node/edge type sets \\
$t_p, t_{tr}$ 
& Primary/trigger node types \\
$\mathcal{T}_{\mathrm{aux}}$ 
& Auxiliary node types \\
$A_{t_a, t_b} \in\{0,1\}^{\left|\mathcal{V}_{t_a}\right| \times\left|\mathcal{V}_{t_b}\right|}$
& Adjacency matrix between types $t_a,t_b$ \\
$\mathcal{Y}$ 
& Class label set \\
$y_t$ 
& Target class \\
$\mathcal{V}_{t_p}, \mathcal{V}^{(p)}$
& Primary-type nodes, poisoned subset \\
$\mathcal{V}_{y_t}, \mathcal{V}_{\neg y_t}$ 
& Primary-type nodes w/ or w/o label $y_t$ \\
$\mathcal{V}_{t_{\mathrm{tr}}}^{(\mathrm{new})}$ 
& Newly added trigger nodes \\
$X^{(\mathrm{new})}, \mathcal{E}^{(\mathrm{new})}$
& New trigger-node features, edges \\
$\widetilde{G}=(\widetilde{\mathcal{V}}, \widetilde{\mathcal{E}}, \widetilde{X})$ 
& Poisoned graph \\
$f_{\theta}$ 
& Classification model \\
$\mathcal{F}(G)$ 
& Allowed modifications \\
$\mathbf{1}(\cdot)$ 
& Indicator function \\
\bottomrule
\end{tabular}
\end{table}

\section{Methodology}
In this section, we introduce the details of HeteroBA, which aims to meet Eq.~(1) to conduct backdoor attacks on heterogeneous graphs. Since directly optimizing the features and connections of the inserted trigger node $v^{(new)}_{t_{t r}}$ to ensure both attack effectiveness and stealthiness is computationally expensive and challenging, HeteroBA decomposes the attack process into two key components, addressing the following two core challenges:
(i) how to generate the features of the inserted trigger node $v^{(new)}_{t_{t r}}$ to enhance their stealthiness and make them less detectable;
(ii) how to construct the connections of the inserted trigger node $v^{(new)}_{t_{t r}}$ to maximize attack effectiveness while maintaining structural consistency with the original graph. To tackle these challenges, HeteroBA consists of two main modules. The Feature Generator is responsible for generating trigger node features by learning the distribution of existing $\mathcal{V}_{t_{t r}}$ nodes. This process ensures that the injected trigger nodes blend seamlessly into the overall feature space, thereby improving stealthiness. The Edge Generator determines how the inserted trigger node $v^{(new)}_{t_{t r}}$ connects to existing nodes. Specifically, $v^{(new)}_{t_{t r}}$ is linked to primary-type nodes $v_{t_p} \in \mathcal{V}_{t_p}$ and auxiliary type nodes $v_{t_a} \in \mathcal{T}_{\text {aux }}$. To establish these connections, HeteroBA introduces two strategies: a clustering-based strategy and an attention-based strategy.

By integrating these two modules, HeteroBA generates a new poisoned graph $\widetilde{G} = (\widetilde{\mathcal{V}}, \widetilde{\mathcal{E}}, \widetilde{X})$, which ensures both high attack effectiveness and strong stealthiness. Fig. \ref{fig:wholeprocess} provides a conceptual illustration of how HeteroBA generates trigger nodes and establishes connections. The pseudocodes of this process are provided in Appendix \ref{appendix_sec: pseudocode}, and the corresponding time complexity analysis is detailed in Appendix \ref{appendix_sec:time complexity analysis}.

\subsection{Feature generator}
In HeteroBA, the Feature Generator is responsible for generating feature embeddings for the inserted trigger nodes $v^{(new)}_{t_{tr}} $, ensuring that they remain indistinguishable from existing nodes in the feature space. To achieve this, we first extract the set of non-target class nodes $\mathcal{V}_{\neg y_t}$ and identify their neighbors connected via edges of type $r_{t_p, t_{tr}}$. We can get the subset of trigger-type nodes $\mathcal{V}_{t_{tr}}$ that are linked to $\mathcal{V}_{\neg y_t}$ and denote this subset as $\mathcal{V}^{\prime}_{t_{tr}}$, which is used as the basis for feature generation by two strategies, one for continuous features and one for binary features.

\subsubsection{Continuous Feature Generation}
For each feature dimension \( j \in \{1,2,\ldots,d\} \) in \( X'_{t_{tr}} \), we employ Kernel Density Estimation (KDE)~\cite{KDE} to approximate the underlying probability distribution:
\begin{equation}
\label{eq:kde}
\hat{f}(x) \;=\; \frac{1}{mh} \sum_{k=1}^{m} K\!\Bigl(\frac{x - x_k}{h}\Bigr),
\end{equation}
where $K(\cdot)$ is the kernel function (commonly Gaussian), $h$ is the bandwidth, and $\left\{x_k\right\}_{k=1}^m$ represents the feature values of dimension $j$ for all nodes in $\mathcal{V}_{t_{t r}}^{\prime}$, with $m=\left|\mathcal{V}_{t_{t r}}^{\prime}\right|$ denoting the number of nodes in this subset. Let \( \hat{f}_j \) denote the fitted KDE for the \(j\)-th feature dimension. We then sample the feature values for newly inserted trigger nodes \( \mathcal{V}_{t_{tr}}^{(\mathrm{new})} \) from this estimated distribution:
\begin{equation}
\label{eq:new_feature_sampling}
X^{(\mathrm{new})}(i, j) \;\sim\; \hat{f}_j,\quad 
\forall \; i \;=\;1,\ldots,\bigl|\mathcal{V}_{t_{tr}}^{(\mathrm{new})}\bigr|\text{ and } j=1,\ldots,d.
\end{equation}
This ensures that each dimension of the newly generated features follows the same statistical profile as the existing trigger-type nodes in \(\mathcal{V}'_{t_{tr}}\). Consequently, the resulting feature matrix \( X^{(\mathrm{new})} \) for the inserted nodes seamlessly aligns with the original distribution, thereby enhancing the stealth of the injected triggers.

\subsubsection{Binary Feature Generation}
When node features are binary (e.g., indicating a categorical attribute or the presence/absence of a property), directly applying KDE is not feasible. Instead, we compute the empirical probability of each feature being 1 from the extracted set \(\mathcal{V}'_{t_{tr}}\). Specifically, for each binary feature dimension \(j\in\{1,2,\ldots,d\}\), let
\begin{equation}
\label{eq:binary_probability}
\hat{p}_j \;=\; \frac{1}{m}\,\sum_{k=1}^{m} X'_{t_{tr}}(k, j),
\end{equation}
where \(X'_{t_{tr}}(k, j)\) is the \(j\)-th feature of the \(k\)-th node in \(\mathcal{V}'_{t_{tr}}\), and \(m = |\mathcal{V}'_{t_{tr}}|\). We then generate the binary features for the newly inserted trigger nodes \(\mathcal{V}_{t_{tr}}^{(\mathrm{new})}\) by sampling each dimension \(j\) via a Bernoulli distribution:
\begin{equation}
\small
\label{eq:new_binary_features}
X^{(\mathrm{new})}(i, j) \;\sim\; \mathrm{Bernoulli}\!\bigl(\hat{p}_j\bigr),
\quad \forall \; i \;=\;1,\ldots,\bigl|\mathcal{V}_{t_{tr}}^{(\mathrm{new})}\bigr| \text{ and } j=1,\ldots,d.
\end{equation}
This process ensures that the newly generated binary features maintain the same empirical probabilities as the existing trigger-type nodes, preserving consistency with the original data distribution. By combining both the continuous and binary feature generation strategies, the feature generator can effectively produce feature embeddings for trigger nodes that blend seamlessly into the heterogeneous graph, thus minimizing the risk of detection.

\subsection{Edge Generator}
To determine which existing nodes should connect with the newly inserted trigger nodes, we first use edge type \(r_{t_p, t_{tr}}\) to gather the first-hop trigger-type neighbors \(\mathcal{V}_{t_{tr}}^{(1)} \subseteq \mathcal{V}_{t_{tr}}\) of \(\mathcal{V}_{y_t}\), then collect the second-hop auxiliary-type neighbors \(\mathcal{V}_{\mathrm{aux}}^{(2)} \subseteq \mathcal{V}_{\mathrm{aux}}\) based on \(\mathcal{V}_{t_{tr}}^{(1)}\) via edges \(r_{t_{tr}, t_b}\) for \(t_b \in \mathcal{T}_{\mathrm{aux}}\). The identified second-hop neighbors serve as the backup nodes, which will be connected to the trigger nodes.

For each auxiliary type \(t_b \in \mathcal{T}_{\mathrm{aux}}\), the number of auxiliary-type nodes connected to each trigger node is denoted as \(d_{t_b}\), which corresponds to the average degree of trigger-type nodes on edges of type \(r_{t_{tr}, t_b}\). We next feed the heterogeneous graph \(G\) into a surrogate model \(\mathcal{M}\) to retrieve attention weights \(\alpha_{(u,v)}\) and node embeddings \(\mathbf{z}_v\). Building on these outputs, the attention-based and clustering-based strategies respectively utilize attention values and embeddings to select the top \(d_{t_b}\) most influential nodes in each auxiliary type \(t_b \in \mathcal{T}_{\mathrm{aux}}\), which are then connected to the newly inserted trigger nodes.

In HeteroBA, we select SimpleHGN \citep{SimpleHGN} as \(\mathcal{M}\) for three primary reasons: it directly yields both node embeddings and attention scores, has comparatively fewer parameters than other heterogeneous models, and requires no explicit metapath definition, thereby mitigating potential biases from metapath design. 

\subsubsection{Attention-based strategy}
We leverage the learned attention coefficients \(\alpha_{(u,v)}\) from the surrogate model \(\mathcal{M}\) to quantify the importance of each node in \(\mathcal{V}_{\mathrm{aux}}^{(2)}\) in influencing \(\mathcal{V}_{y_t}\). Specifically, for each target node \(v_{y_t} \in \mathcal{V}_{y_t}\), we consider its first-hop trigger-type neighbors in \(\mathcal{V}_{t_{tr}}^{(1)}\) via edges of type \(r_{t_p, t_{tr}}\), and subsequently aggregate contributions from their second-hop neighbors in \(\mathcal{V}_{\mathrm{aux}}^{(2)}\) through edges \(r_{t_{tr}, t_b}\).

For an auxiliary-type node \(v_{\mathrm{aux}}^{(2)} \in \mathcal{V}_{\mathrm{aux}}^{(2)}\), its influence on \(v_{y_t}\) is determined by the product of the first-layer attention value \(\alpha_{(v_{\mathrm{aux}}^{(2)}, v_{t_{tr}}^{(1)})}\) and the second-layer attention value \(\alpha_{(v_{t_{tr}}^{(1)}, v_{y_t})}\):

\begin{equation}
I\bigl(v_{\mathrm{aux}}^{(2)}, v_{y_t}\bigr)
=
\sum_{v_{t_{tr}}^{(1)} \in \mathcal{V}_{t_{tr}}^{(1)}(v_{y_t})}
\alpha_{(v_{\mathrm{aux}}^{(2)}, v_{t_{tr}}^{(1)})}
\cdot
\alpha_{(v_{t_{tr}}^{(1)}, v_{y_t})}.
\end{equation}
where \(\mathcal{V}_{t_{tr}}^{(1)}(v_{y_t})\) represents the set of first-hop trigger-type neighbors of \(v_{y_t}\).

Summing over all \(v_{y_t} \in \mathcal{V}_{y_t}\), we compute the total importance score for each auxiliary-type node:

\begin{equation}
I\bigl(v_{\mathrm{aux}}^{(2)}\bigr)
=
\sum_{v_{y_t} \in \mathcal{V}_{y_t}}
I\bigl(v_{\mathrm{aux}}^{(2)}, v_{y_t}\bigr).
\end{equation}

Nodes in \(\mathcal{V}_{\mathrm{aux}}^{(2)}\) with higher \(I(v_{\mathrm{aux}}^{(2)})\) scores are ranked in descending order, and the top-ranked nodes in each auxiliary type are selected to connect with the newly inserted trigger nodes. This ensures that adversarial information is efficiently propagated toward \(\mathcal{V}_{y_t}\), thereby enhancing the backdoor attack effectiveness while maintaining the structural consistency of the heterogeneous graph.

\subsubsection{Clustering-based Strategy}
In another strategy, we leverage the node embeddings obtained from the surrogate model \(\mathcal{M}\) to identify structurally and semantically cohesive nodes within \(\mathcal{V}_{\mathrm{aux}}^{(2)}\). Specifically, we extract the embedding representations of all second-hop auxiliary-type neighbors \(\mathcal{V}_{\mathrm{aux}}^{(2)}\) and employ a clustering-based selection strategy to determine which nodes should connect to the newly inserted trigger nodes.

Given the embedding matrix \(\mathbf{Z} \in \mathbb{R}^{|\mathcal{V}_{\mathrm{aux}}^{(2)}| \times d}\), where each row \(\mathbf{z}_{i}\) corresponds to the \(d\)-dimensional embedding of node \(v_{\mathrm{aux}, i}^{(2)} \in \mathcal{V}_{\mathrm{aux}}^{(2)}\), we first compute the pairwise cosine similarity matrix:

\begin{equation}
S = \mathbf{Z} \cdot \mathbf{Z}^{T},
\end{equation}
where the entry \( S_{ij} \) represents the cosine similarity between nodes \( v_{\mathrm{aux}, i}^{(2)} \) and \( v_{\mathrm{aux}, j}^{(2)} \). To ensure that self-similarity does not dominate the selection process, we set the diagonal elements of \( S \) to zero:

\begin{equation}
S_{ii} = 0, \quad \forall i \in \{1,2,\dots, |\mathcal{V}_{\mathrm{aux}}^{(2)}|\}.
\end{equation}

For each node \( v_{\mathrm{aux}, i}^{(2)} \), we compute its average similarity score, defined as:

\begin{equation}
I(v_{\mathrm{aux}, i}^{(2)}) = \frac{1}{|\mathcal{V}_{\mathrm{aux}}^{(2)}| - 1} \sum_{j \neq i} S_{ij}.
\end{equation}

Nodes in \(\mathcal{V}_{\mathrm{aux}}^{(2)}\) with higher \( I(v_{\mathrm{aux}, i}^{(2)}) \) scores exhibit stronger embedding similarities to other nodes within the same type, indicating their centrality within structurally cohesive regions. We rank all nodes in descending order based on \( I(v_{\mathrm{aux}, i}^{(2)}) \) and select the top-ranked nodes within each auxiliary type to connect with the newly inserted trigger nodes. 

By enforcing connections with the most clustered and semantically aligned nodes, this strategy ensures that the inserted trigger nodes integrate seamlessly into the graph structure, thereby improving stealthiness while maintaining attack effectiveness.

\section{Experiments}
In this section, we evaluate our proposed method on multiple benchmark datasets to investigate the following research questions:
\begin{enumerate}[label=\textbf{RQ\arabic*:}]
    \item How effective is the attack?
    \item How is the stealthiness?
    \item Do the two edge-generation strategies (e.g., attention-based or clustering-based) indeed improve attack performance ?
    \item What is the relationship between the poison rate and attack effectiveness?
    \item Why do these particular auxiliary-type nodes effectively facilitate backdoor infiltration in a heterogeneous graph, and what crucial role do they play in the attack success?
\end{enumerate}

\subsection{Experimental Settings}
\subsubsection{Datasets}
We evaluate our method on three real-world heterogeneous datasets: DBLP, ACM, and IMDB\citep{MAGNN}. DBLP consists of four entity types (authors, papers, terms, conferences), with authors being categorized into three research areas (database, data mining, artificial intelligence). ACM includes papers from KDD, SIGMOD, SIGCOMM, MobiCOMM, and VLDB, being categorized into three fields (database, wireless communication, data mining). IMDB contains movies, keywords, actors, and directors, with movies being classified into action, comedy, and drama. The statistics of these datasets are shown in Table~\ref{tab:dataset statistics}.

\begin{table}[ht]
\centering
\small
\caption{Dataset Statistics}
\label{tab:dataset statistics}
\resizebox{\linewidth}{!}{ 
\begin{tabular}{lccccc}
\toprule
\textbf{Dataset} & \textbf{\#Node Types} & \textbf{\#Edge Types} & \textbf{\#Nodes} & \textbf{\#Edges} & \textbf{Primary Type} \\
\midrule
ACM   & 3 & 4 & 11252  & 34864  & paper  \\
IMDB  & 3 & 4 & 11616  & 34212  & movie \\
DBLP  & 4 & 6 & 26198  & 242142 & author  \\
\bottomrule
\end{tabular}
}
\end{table}

\subsubsection{Train settings}

We conduct experiments using HAN \citep{HAN}, HGT \citep{HGT}, and SimpleHGN \citep{SimpleHGN} as victim models, ensuring a fair comparison of backdoor attack performance under the same training and evaluation conditions. The dataset is divided into training, testing, and validation sets. The training set comprises 70\% of the primary-type nodes $\mathcal{V}_{t_p}$, including both clean and poisoned nodes. Specifically, the poisoned training set (Poison Trainset) accounts for 5\% of $\mathcal{V}_{t_p}$, serving as the injected trigger nodes to facilitate backdoor activation. The testing set constitutes 20\% of $\mathcal{V}_{t_p}$, within which the poisoned testing set (Poison Testset) also accounts for 5\%, allowing us to evaluate the attack's effectiveness during inference. The remaining 10\% is allocated to the validation set, which is used for hyperparameter tuning and early stopping. The training parameters are provided in Appendix \ref{appendix_sec:other training parameters}.

\subsubsection{Compared Methods}
Since our work is the first to explore backdoor attacks on heterogeneous graphs, we adapt existing backdoor attack methods originally designed for homogeneous graphs, namely UGBA \citep{UGBA} and CGBA \citep{CGBA}, and modify them to be compatible with heterogeneous graphs, using the adapted versions as our baselines.

For UGBA, to ensure its applicability to heterogeneous graphs, we first convert the heterogeneous graph into a homogeneous graph. Following UGBA's bi-level optimization strategy, we generate and inject adversarial graph structures. After completing the attack in the homogeneous setting, we convert the graph back into its heterogeneous form. During this process, the newly inserted nodes and edges introduced by UGBA are assigned random node types and edge types to conform to the heterogeneous graph schema.

For CGBA, as it does not involve structural modifications, we directly adapt its feature perturbation strategy to heterogeneous graphs by applying it to the nodes in $\mathcal{V}^{(p)}$, ensuring its effectiveness in this setting.

\subsubsection{Evaluation Metrics}

The \textit{Attack Success Rate (ASR)} \citep{UGBA} measures the probability that the backdoored model \( f_b \) misclassifies a sample embedded with a trigger \( g_t \) into the target class \( y_t \). Formally, ASR is defined as:

\begin{equation}
ASR = \frac{\sum_{i=1}^{n} \mathbf{1}(f_b(v_i) = y_t)}{n}
\end{equation}
where \( n \) denotes the number of poisoned test samples, and \( \mathbf{1}(\cdot) \) represents the indicator function. A higher ASR indicates a more effective backdoor attack.

The \textit{Clean Accuracy Drop (CAD)} \citep{CGBA} quantifies the degradation in classification accuracy of the backdoored model \( f_b \) on clean samples compared to the clean model \( f_c \). It is defined as:

\begin{equation}
CAD = Acc_{f_c}(\text{Clean}) - Acc_{f_b}(\text{Clean})
\end{equation}
where \( Acc_{f_c}(\text{Clean}) \) and \( Acc_{f_b}(\text{Clean}) \) denote the prediction accuracies of the clean model \( f_c \) and the backdoored model \( f_b \) on clean samples, respectively. A lower CAD indicates that the backdoor attack preserves the original model’s performance on clean data.

We introduce a method to calculate the \textit{Stealthiness Score} for node injection-based methods in heterogeneous graphs, providing a quantitative measure for evaluating the concealment of injected nodes. This score assesses the similarity between injected trigger nodes and clean nodes in both feature and structural aspects. Given the original graph \(G\) and the poisoned graph \(\widetilde{G}\), we compute the feature similarity \(\text{Sim}_{\text{feat}}\) and structural similarity \(\text{Sim}_{\text{struct}}\), then combine them to obtain the final score.

Feature similarity measures how closely the injected nodes' feature distribution matches that of clean nodes. Let \(\mathcal{V}_{t_{tr}}^{(new)}\) be the set of newly injected trigger nodes and \(\mathcal{V}_{t_{tr}}^{(clean)}\) the clean nodes of the same type. For each feature dimension \(i\), we compute the Wasserstein distance \(\text{WD}_i\) between the feature distributions of these two sets and define the average Wasserstein distance as:
\begin{equation}
\overline{\text{WD}} = \frac{1}{d} \sum_{i=1}^{d} \text{WD}_{i}.
\end{equation}
Feature similarity is then given by:
\begin{equation}
\text{Sim}_{\text{feat}} = \frac{1}{1+\overline{\text{WD}}}.
\end{equation}

Structural similarity evaluates the degree consistency between injected and clean nodes. Let \(\bar{d}_{trg}\) and \(\bar{d}_{clean}\) be the average degrees of injected and clean nodes, respectively. The degree difference is defined as:
\begin{equation}
\Delta d = |\bar{d}_{trg} - \bar{d}_{clean}|,
\end{equation}
and the structural similarity is computed as:
\begin{equation}
\text{Sim}_{\text{struct}} = \frac{1}{1+\Delta d}.
\end{equation}

The final stealthiness score is a weighted sum of both components:
\begin{equation}
\text{Stealthiness}(G, \widetilde{G}) = w_1 \cdot \text{Sim}_{\text{feat}} + w_2 \cdot \text{Sim}_{\text{struct}},
\end{equation}
where \(w_1\) and \(w_2\) are weighting factors (default to 0.5). A higher score indicates that the injected nodes blend more naturally into the graph, enhancing attack stealthiness. Unlike previous works, which qualitatively discuss stealthiness, our proposed \textit{Stealthiness Score} provides a quantitative measure, enabling a more precise evaluation of attack concealment.

\subsection{Experiment result}
\begin{table*}[htbp]
\centering
\scriptsize
\caption{Attack effectiveness on three datasets (ACM, DBLP, IMDB).}
\label{tab:best-triggers}
\begin{tabular}{c|c|c|c|ccccc|ccccc}
\hline
\multirow{2}{*}{\textbf{Dataset}} 
 & \multirow{2}{*}{\textbf{Victim Model}} 
 & \multirow{2}{*}{\textbf{Class}} 
 & \multirow{2}{*}{\textbf{Trigger}} 
 & \multicolumn{5}{c|}{\textbf{ASR}} 
 & \multicolumn{5}{c}{\textbf{CAD}} \\
\cline{5-14}
 &  &  & 
 & \textbf{HeteroBA-C} & \textbf{HeteroBA-A} & \textbf{HeteroBA-R} & \textbf{CGBA} & \textbf{UGBA} 
 & \textbf{HeteroBA-C} & \textbf{HeteroBA-A} & \textbf{HeteroBA-R} & \textbf{CGBA} & \textbf{UGBA} \\
\hline
\multirow{9}{*}{\textbf{ACM}}
 & \multirow{3}{*}{HAN}
   & 0 & \multirow{3}{*}{author}
     & \textbf{0.9983} & 0.3748 & 0.3416 & 0.9420 & 0.0664
     & \textbf{-0.0033} & -0.0028 & $-0.0558^\dagger$ & 0.0160 & 0.0099 \\
 &  & 1 & 
     & \textbf{1.0000} & 0.7463 & 0.8275 & 0.5970 & 0.9783
     & \textbf{-0.0005} & 0.0480 & $-0.0602^\dagger$ & 0.1109 & 0.0149 \\
 &  & 2 &
     & \textbf{1.0000} & 0.7861 & 0.5224 & 0.8126 & 0.0498
     & -0.0132 & -0.0193 & $-0.0480^\dagger$ & \textbf{-0.0375} & 0.0149 \\
\cline{2-14}
 & \multirow{3}{*}{HGT}
   & 0 & \multirow{3}{*}{author}
     & \textbf{1.0000} & \textbf{1.0000} & 0.9751 & 0.9436 & 0.9867
     & -0.0033 & \textbf{-0.0044} & 0.0226 & -0.0022 & -0.0038 \\
 &  & 1 &
     & 0.9569 & 0.9469 & $0.9851^\dagger$ & 0.8905 & \textbf{0.9851}
     & \textbf{-0.0061} & -0.0028 & -0.0017 & 0.0122 & \textbf{-0.0061} \\
 &  & 2 &
     & \textbf{1.0000} & \textbf{1.0000} & 0.7977 & 0.9005 & 0.9469
     & 0.0050 & \textbf{-0.0027} & 0.0138 & 0.0083 & 0.0006 \\
\cline{2-14}
 & \multirow{3}{*}{SimpleHGN}
   & 0 & \multirow{3}{*}{author}
     & 0.9967 & \textbf{1.0000} & 0.9536 & 0.9602 & \textbf{1.0000}
     & -0.0027 & 0.0000 & 0.0099 & -0.0022 & \textbf{-0.0254} \\
 &  & 1 &
     & 0.9950 & \textbf{1.0000} & 0.9967 & 0.9303 & \textbf{1.0000}
     & 0.0028 & 0.0033 & -0.0011 & 0.0000 & \textbf{-0.0750} \\
 &  & 2 &
     & \textbf{1.0000} & \textbf{1.0000} & 0.6965 & 0.9038 & \textbf{1.0000}
     & 0.0011 & 0.0011 & 0.0111 & 0.0033 & \textbf{-0.0695} \\
\hline
\multirow{9}{*}{\textbf{DBLP}}
 & \multirow{3}{*}{HAN}
   & 0 & \multirow{3}{*}{paper}
     & 0.7849 & 0.7783 & 0.2167 & \textbf{0.8993} & 0.0673
     & 0.0110 & 0.0104 & $0.0005^\dagger$ & \textbf{0.0038} & 0.0094 \\
 &  & 1 &
     & 0.6043 & 0.6716 & 0.1576 & \textbf{0.9422} & 0.0541
     & 0.0214 & 0.0411 & 0.0033 & \textbf{0.0022} & 0.0027 \\
 &  & 2 &
     & 0.6749 & 0.5534 & 0.1855 & \textbf{0.9142} & 0.0312
     & 0.0143 & 0.0011 & 0.0104 & \textbf{0.0088} & -0.0115 \\
\cline{2-14}
 & \multirow{3}{*}{HGT}
   & 0 & \multirow{3}{*}{paper}
     & \textbf{0.9343} & 0.9130 & $0.9950^\dagger$ & 0.9175 & 0.0788
     & 0.0137 & 0.0165 & 0.0148 & \textbf{-0.0027} & 0.0088 \\
 &  & 1 &
     & 0.7980 & \textbf{0.9967} & 0.7537 & 0.9505 & 0.1117
     & \textbf{0.0099} & 0.0137 & 0.0115 & 0.0016 & 0.0110 \\
 &  & 2 &
     & 0.8588 & 0.8867 & 0.6650 & \textbf{0.9439} & 0.1790
     & 0.0137 & 0.0131 & 0.0181 & \textbf{0.0077} & 0.0192 \\
\cline{2-14}
 & \multirow{3}{*}{SimpleHGN}
   & 0 & \multirow{3}{*}{paper}
     & \textbf{1.0000} & \textbf{1.0000} & 0.9984 & 0.9224 & 0.0263
     & \textbf{0.0016} & 0.0033 & 0.0082 & 0.0077 & 0.0044 \\
 &  & 1 &
     & 0.9754 & \textbf{1.0000} & 0.6897 & 0.9538 & 0.1347
     & 0.0055 & 0.0033 & $0.0027^\dagger$ & \textbf{-0.0016} & 0.0066 \\
 &  & 2 &
     & 0.7307 & 0.9294 & 0.7093 & \textbf{0.9422} & 0.1084
     & \textbf{-0.0060} & 0.0071 & 0.0077 & 0.0011 & -0.0044 \\
\hline
\multirow{9}{*}{\textbf{IMDB}}
 & \multirow{3}{*}{HAN}
   & 0 & \multirow{3}{*}{director}
     & \textbf{0.9953} & 0.8006 & 0.6791 & 0.5618 & 0.2087
     & 0.0307 & 0.0089 & 0.0265 & \textbf{0.0037} & 0.0364 \\
 &  & 1 &
     & \textbf{0.9984} & 0.8473 & 0.8458 & 0.4523 & 0.2991
     & -0.0031 & \textbf{-0.0192} & -0.0094 & -0.0119 & 0.0037 \\
 &  & 2 &
     & \textbf{1.0000} & 0.9174 & 0.9003 & 0.4992 & 0.3582
     & 0.0068 & \textbf{-0.0234} & -0.0068 & 0.0010 & 0.0067 \\
\cline{2-14}
 & \multirow{3}{*}{HGT}
   & 0 & \multirow{3}{*}{director}
     & 0.8473 & \textbf{0.9237} & 0.7975 & 0.4851 & 0.5109
     & 0.0036 & 0.0062 & 0.0021 & \textbf{-0.0104} & 0.0291 \\
 &  & 1 &
     & \textbf{0.9299} & 0.9283 & 0.8878 & 0.4147 & 0.7757
     & 0.0182 & 0.0234 & $-0.0146^\dagger$ & \textbf{0.0130} & 0.0026 \\
 &  & 2 &
     & 0.8894 & \textbf{0.9377} & 0.8193 & 0.4523 & 0.6807
     & 0.0026 & \textbf{-0.0026} & $-0.0099^\dagger$ & -0.0015 & 0.0182 \\
\cline{2-14}
 & \multirow{3}{*}{SimpleHGN}
   & 0 & \multirow{3}{*}{director}
     & 0.9533 & \textbf{0.9813} & 0.7679 & 0.3881 & 0.8443
     & -0.0047 & 0.0021 & 0.0015 & \textbf{-0.0244} & 0.0005 \\
 &  & 1 &
     & 0.9502 & 0.9564 & 0.9486 & 0.3850 & \textbf{0.9595}
     & 0.0047 & 0.0109 & 0.0052 & \textbf{-0.0130} & 0.0291 \\
 &  & 2 &
     & \textbf{0.9720} & 0.9642 & 0.8255 & 0.3474 & 0.9330
     & \textbf{-0.0052} & 0.0099 & $-0.0166^\dagger$ & 0.0156 & 0.0078 \\
\hline
\end{tabular}
\end{table*}
\subsubsection{Attack effectiveness}
To answer \textbf{RQ1}, we evaluate HeteroBA’s effectiveness against UGBA and CGBA on three heterogeneous graph datasets. Each experiment was repeated three times, with averaged results reported. Table \ref{tab:best-triggers} shows key results, while Table \ref{tab:appendix-triggers} in appendix provides additional data. \textbf{Bold} highlights the best ASR values among HeteroBA-C, HeteroBA-A, UGBA, and CGBA.

HeteroBA consistently achieves high Attack Success Rates (ASR), outperforming UGBA and CGBA across datasets and models. For example, on ACM, HeteroBA-A achieves an ASR of 1.0000 on HAN and HGT, while UGBA reaches only 0.0664 on HAN (class 0). On DBLP, HeteroBA-C achieves 1.0000 ASR on SimpleHGN (class 0), surpassing UGBA’s 0.0263. On IMDB, HeteroBA-A achieves 0.9813 ASR on SimpleHGN (class 0), outperforming CGBA (0.3881) and UGBA (0.8443). These results highlight HeteroBA’s superior attack effectiveness.

Despite high ASR, HeteroBA introduces minimal classification accuracy degradation (CAD), often close to zero or negative, indicating little impact on clean data. Although UGBA achieves slightly lower CAD in some cases (e.g., DBLP, class 1, SimpleHGN), the difference is negligible. Overall, HeteroBA maintains clean data performance while achieving superior attack effectiveness.


\subsubsection{Stealthiness analysis}
As for \textbf{RQ2}, we compare the Stealthiness Score of HeteroBA and UGBA across different datasets and trigger settings, as shown in Table \ref{tab:stealthiness}. CGBA is not included in this comparison since it only perturbs node features without modifying the graph structure, making the Stealthiness Score inapplicable to it.

Results show that HeteroBA consistently outperforms UGBA in stealthiness, indicating that its injected trigger nodes are less detectable. For instance, in ACM with author as the trigger, HeteroBA achieves significantly higher scores across all classes (up to 0.8603), while UGBA remains below 0.2251. In DBLP, where paper is used as the trigger, HeteroBA maintains an advantage, achieving scores around 0.6201 compared to UGBA’s 0.4841. Similarly, in IMDB, when using director as the trigger, HeteroBA achieves a stealthiness score of up to 0.6865, outperforming UGBA's 0.6531. These results confirm that HeteroBA effectively integrates trigger nodes into the original graph through sampled feature and edge generation strategies, improving attack stealthiness.
\begin{table}[htbp]
\centering
\caption{Stealthiness Score of HeteroBA and UGBA.}
\label{tab:stealthiness}
\small
\begin{tabular}{c|c|c|c|c}
\hline
\textbf{Dataset} & \textbf{Class} & \textbf{Trigger} & \textbf{HeteroBA} & \textbf{UGBA} \\
\hline
\multirow{6}{*}{ACM} 
& 0 & \multirow{3}{*}{author} & \textbf{0.7715} & 0.2109 \\ \cline{2-2}
& 1 & & \textbf{0.8603} & 0.2251 \\ \cline{2-2}
& 2 & & \textbf{0.7226} & 0.1899 \\ \cline{2-3} \cline{4-5}
& 0 & \multirow{3}{*}{field} & \textbf{0.3937} & 0.0342 \\ \cline{2-2} 
& 1 & & \textbf{0.3815} & 0.0342 \\ \cline{2-2}
& 2 & & \textbf{0.3539} & 0.0294 \\ \hline
\multirow{3}{*}{DBLP} 
& 0 & \multirow{3}{*}{paper} & \textbf{0.6161} & 0.5063 \\ \cline{2-2}
& 1 & & \textbf{0.6201} & 0.4841 \\ \cline{2-2}
& 2 & & \textbf{0.6179} & 0.4963 \\ \hline
\multirow{6}{*}{IMDB} 
& 0 & \multirow{3}{*}{director} & \textbf{0.6567} & 0.6492 \\ \cline{2-2}
& 1 & & \textbf{0.6711} & 0.6531 \\ \cline{2-2}
& 2 & & \textbf{0.6865} & 0.6297 \\ \cline{2-5}
& 0 & \multirow{3}{*}{actor} & \textbf{0.7663} & 0.6304 \\ \cline{2-2}
& 1 & & \textbf{0.7302} & 0.6252 \\ \cline{2-2}
& 2 & & \textbf{0.7874} & 0.6090 \\ \hline
\end{tabular}
\end{table}

\begin{figure*}[htbp]
    \centering
    \begin{subfigure}{0.25\textwidth}
        \includegraphics[width=\textwidth]{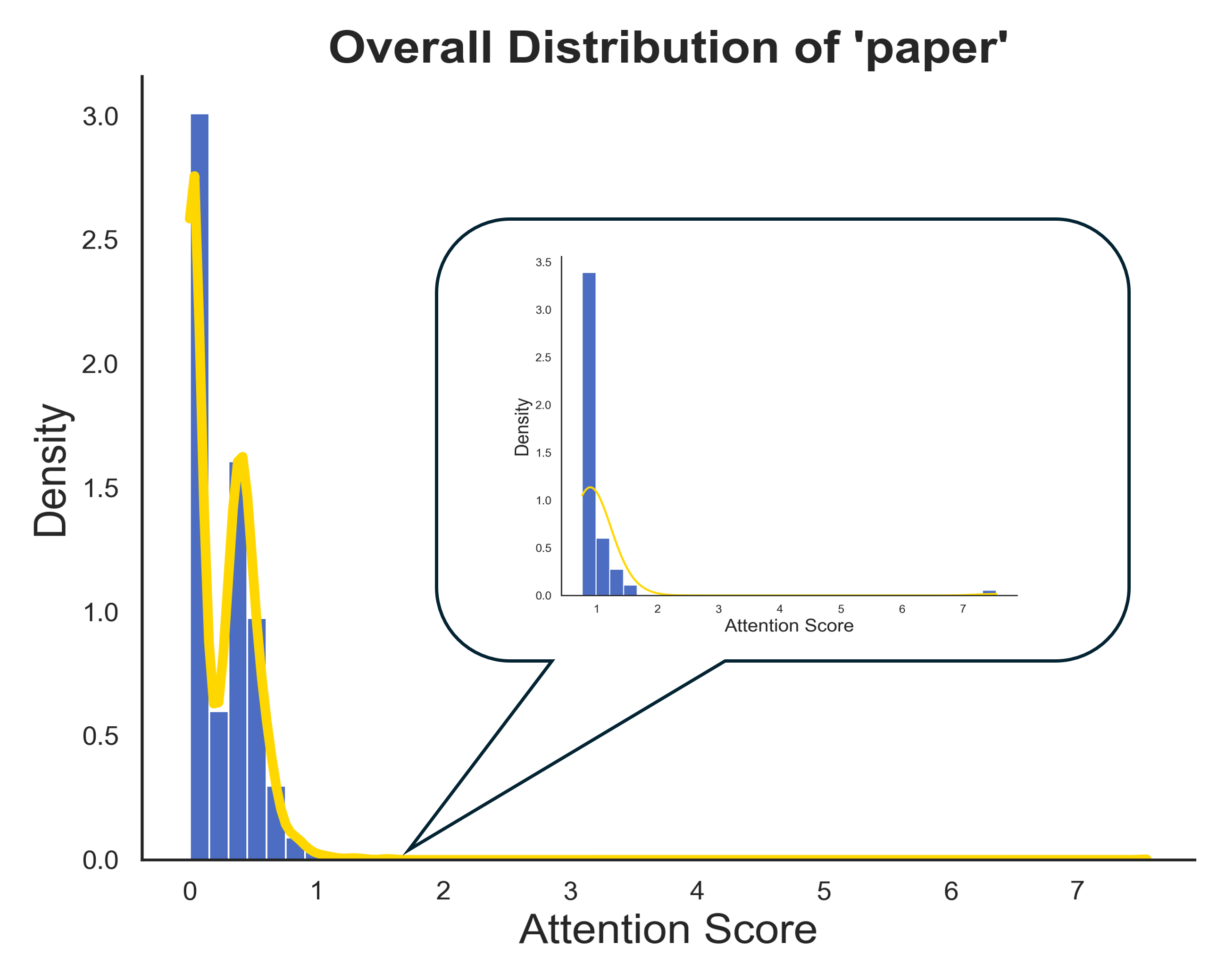}
        \caption{\small ACM Paper Attn. Dist.}
        \label{fig:subfig2}
    \end{subfigure}
     \hspace{0.05\textwidth} 
    \begin{subfigure}{0.25\textwidth}
        \includegraphics[width=\textwidth]{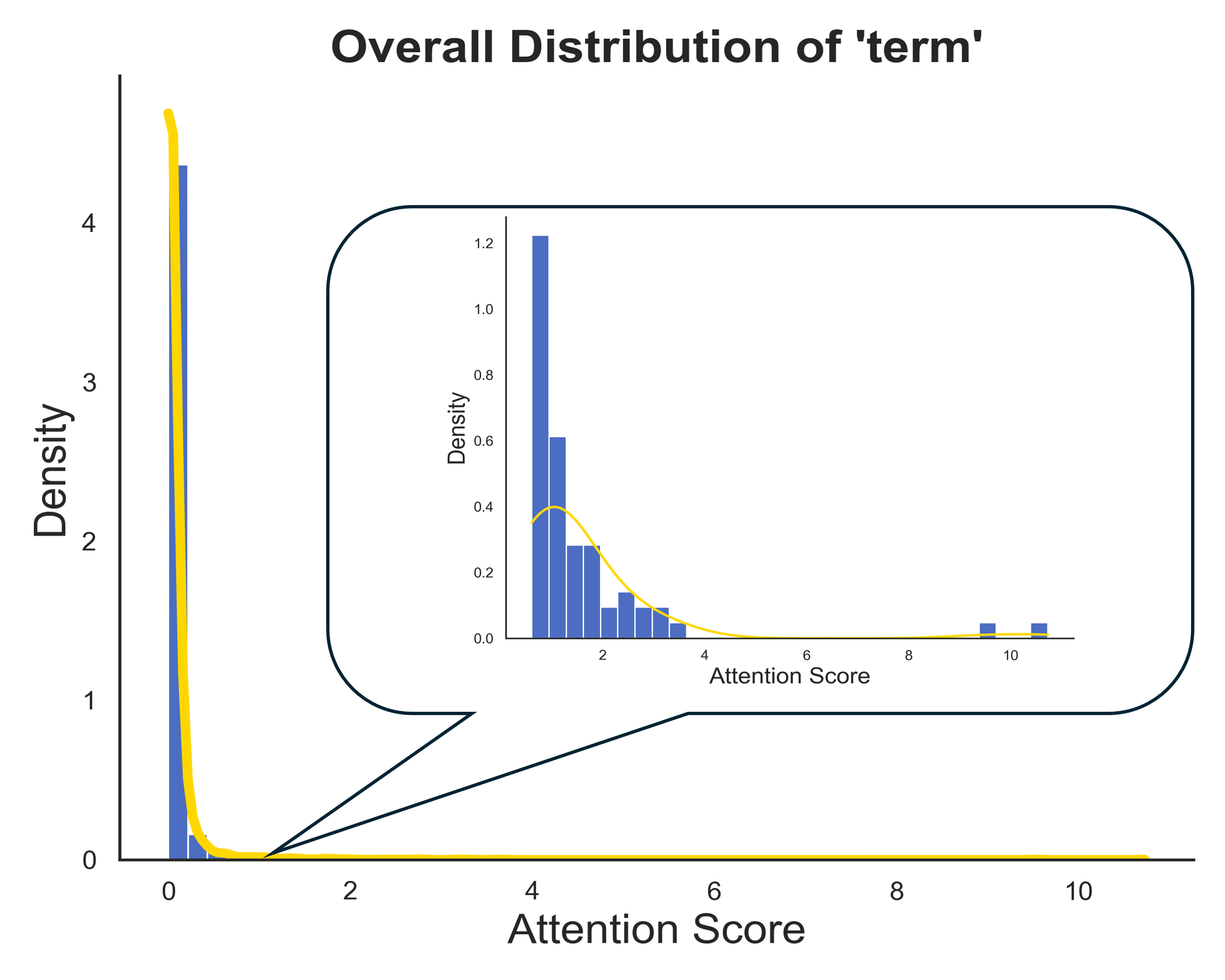}
        \caption{\small DBLP Term Attn. Dist.}
        \label{fig:subfig4}
    \end{subfigure}
     \hspace{0.05\textwidth} 
    \begin{subfigure}{0.25\textwidth}
        \includegraphics[width=\textwidth]{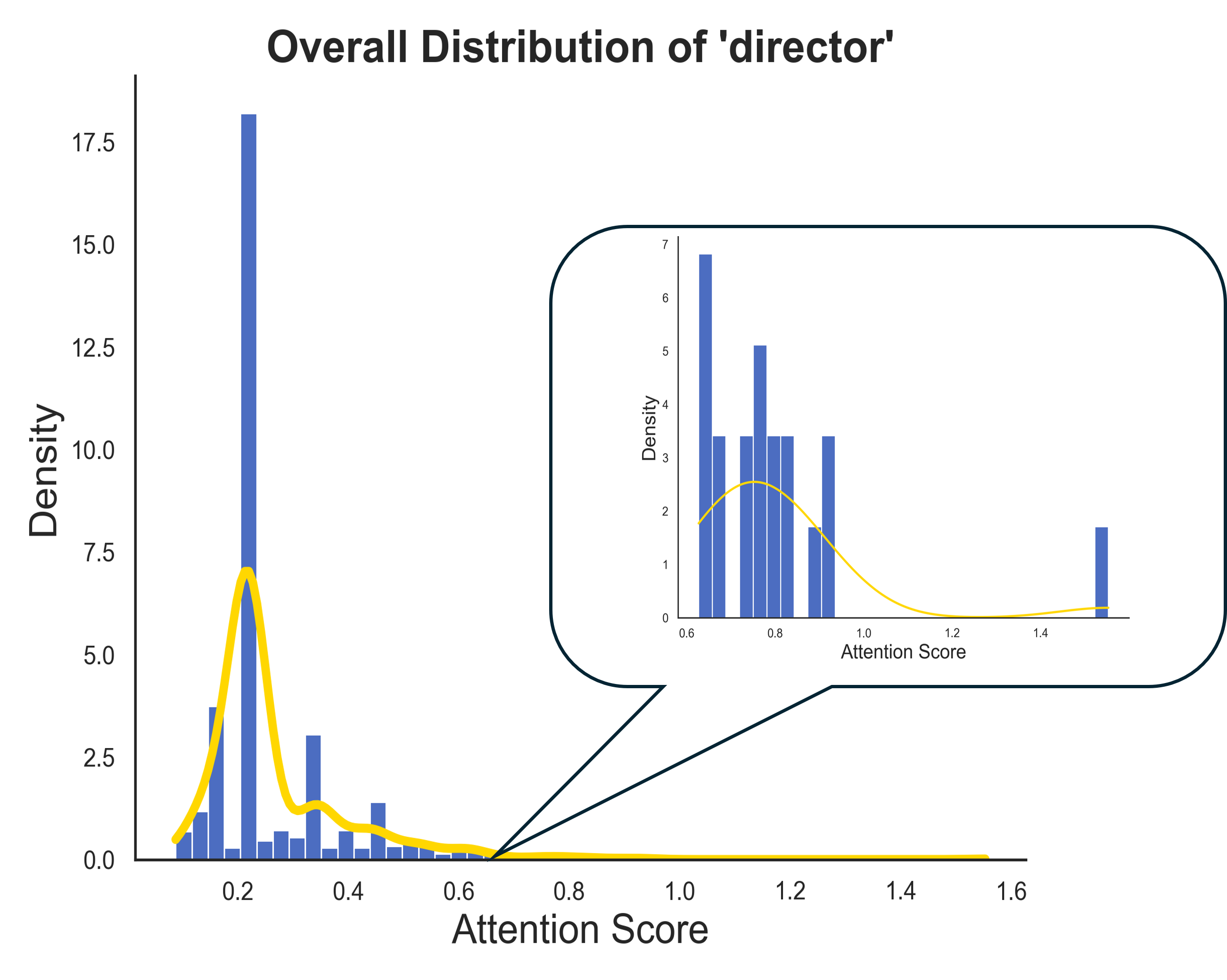}
        \caption{\small IMDB Director Attn. Dist.}
        \label{fig:subfig6}
    \end{subfigure}
    \par\medskip
    \begin{subfigure}{0.25\textwidth}
        \includegraphics[width=\textwidth]{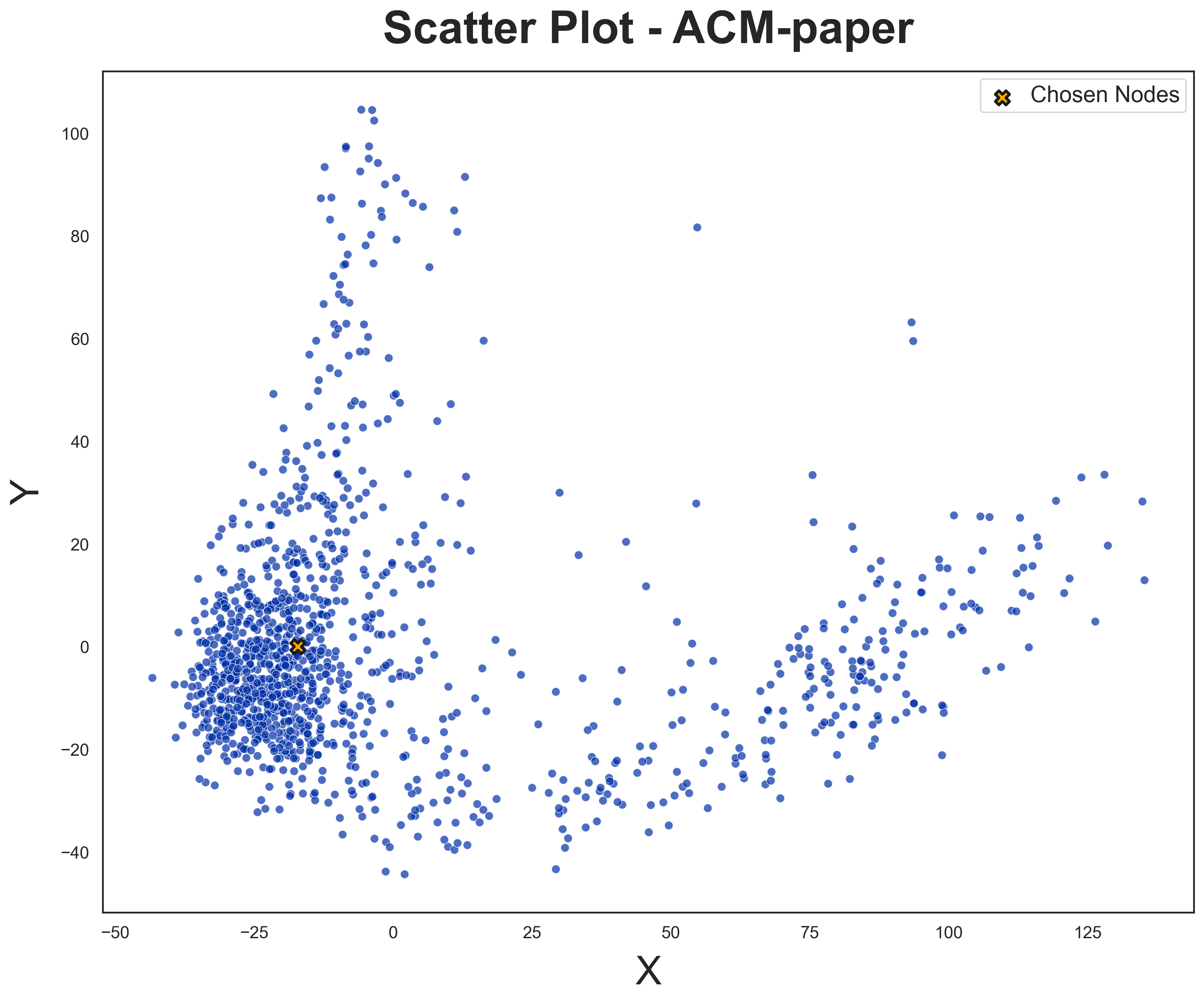}
        \caption{ACM Paper Clustering}
        \label{fig:subfig7}
    \end{subfigure}
     \hspace{0.05\textwidth} 
    \begin{subfigure}{0.25\textwidth}
        \includegraphics[width=\textwidth]{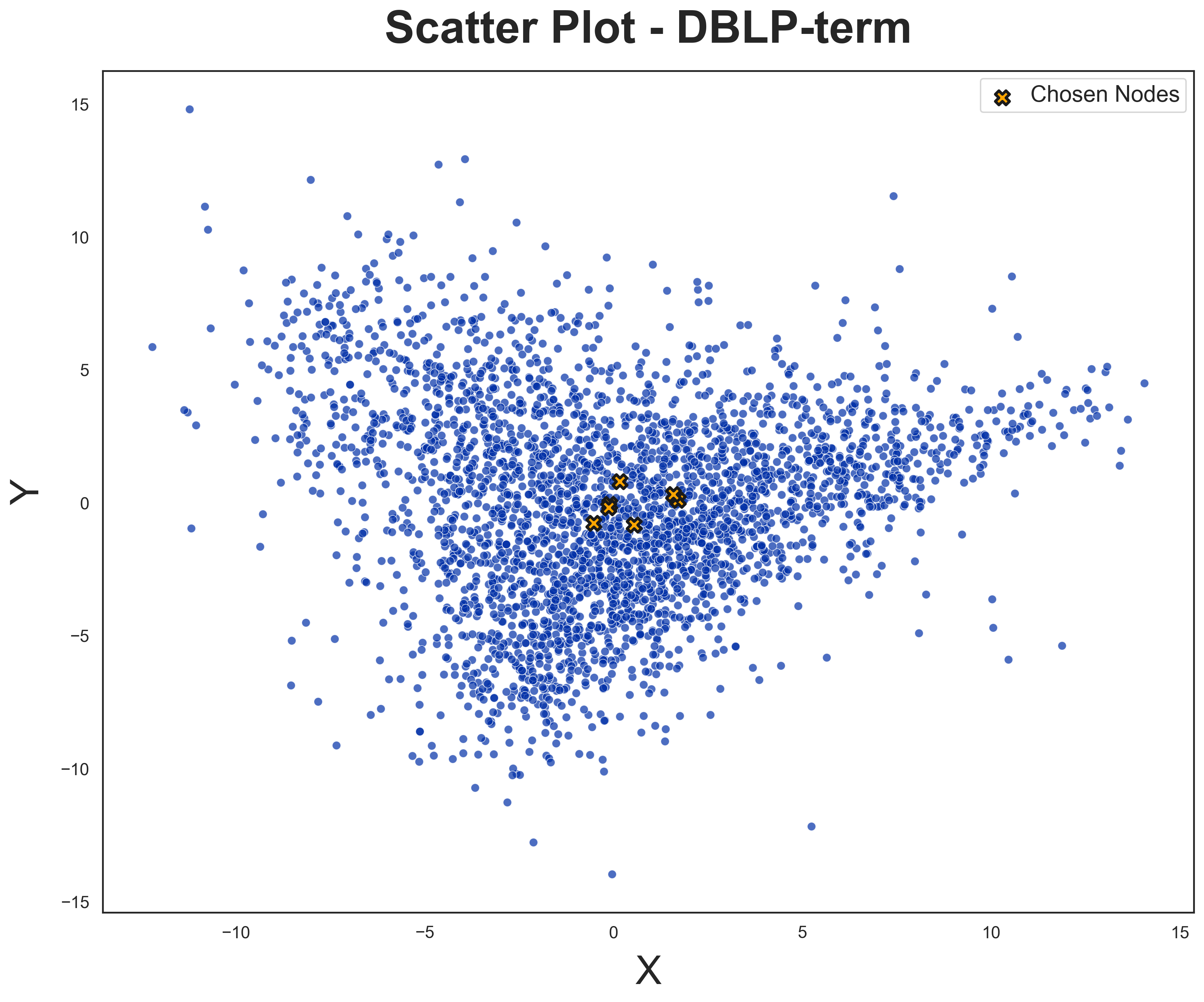}
        \caption{DBLP Term Clustering}
        \label{fig:subfig8}
    \end{subfigure}
     \hspace{0.05\textwidth} 
    \begin{subfigure}{0.25\textwidth}
        \includegraphics[width=\textwidth]{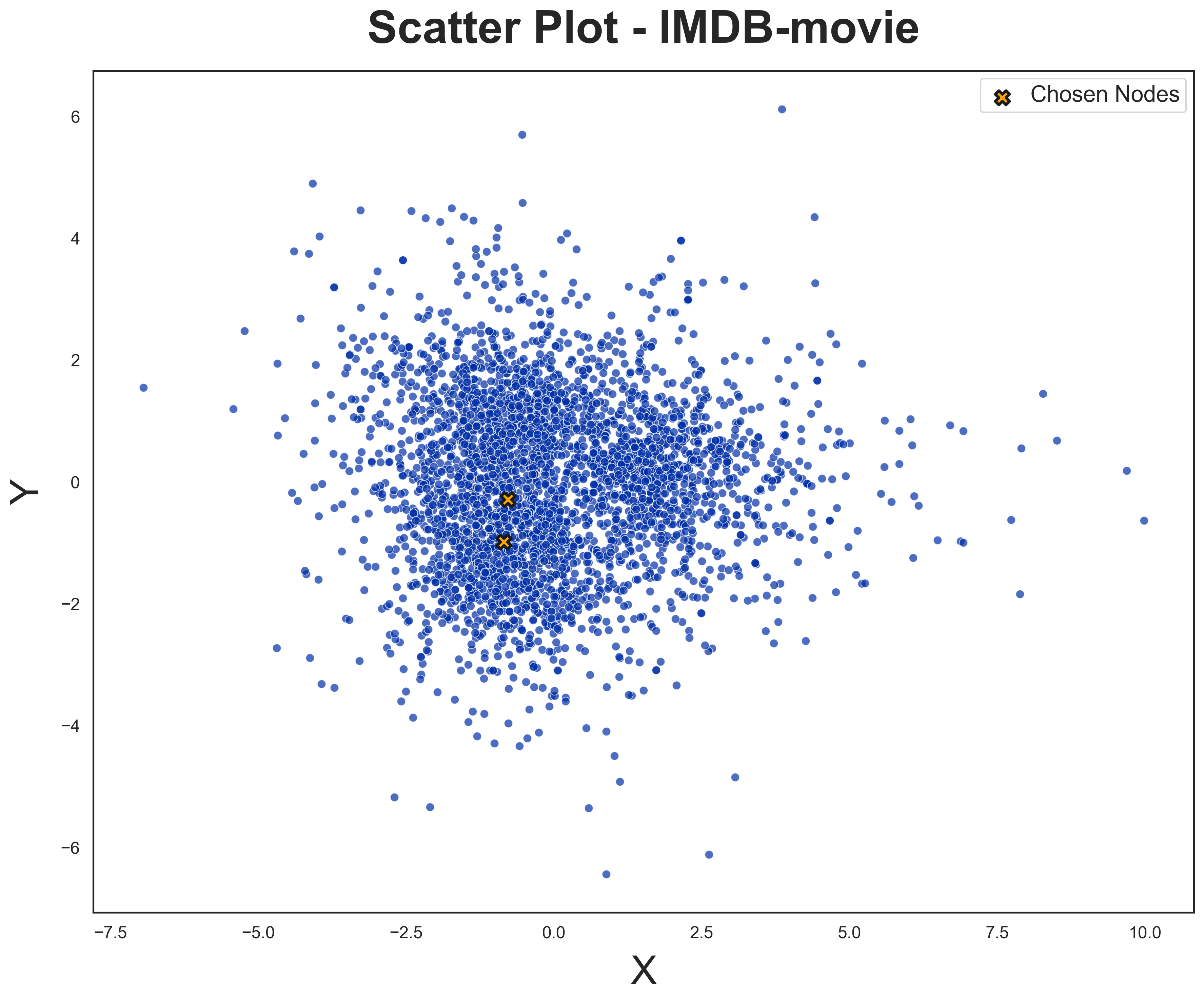}
        \caption{IMDB Movie Clustering}
        \label{fig:subfig9}
    \end{subfigure}
    
    \caption{Attention distribution and Embedding Clustering}
    \label{fig:2x3grid}
\end{figure*}

\subsubsection{Ablation study}
To answer \textbf{RQ3}, we conducted an ablation study by replacing the edge-generation strategies in HeteroBA with a random connection strategy, denoted as HeteroBA-R, to evaluate the impact of the Cluster-based strategy (HeteroBA-C) and the Attention-based strategy (HeteroBA-A). If HeteroBA-R outperforms other methods, we mark it with $\dagger$. The results in Table \ref{tab:best-triggers} show that HeteroBA-R consistently exhibits a significant drop in Attack Success Rate (ASR) across different datasets. For example, in the ACM dataset with author as the trigger, HeteroBA-C achieves an ASR of 1.0000 on HGT (class 2), while HeteroBA-R only reaches 0.7977. Similar trends are observed in DBLP and IMDB, indicating that structured edge-generation strategies help improve ASR.

In terms of Clean Accuracy Drop (CAD), HeteroBA-R does not show a significant advantage. Although CAD is slightly lower in some cases (e.g., in the ACM dataset, HeteroBA-R’s CAD on HAN (class 1) is -0.0602, lower than HeteroBA-C’s -0.0005), the difference is minimal. This suggests that HeteroBA-C and HeteroBA-A can enhance ASR without significantly affecting clean data performance.

In summary, Cluster-based and Attention-based edge-generation strategies are crucial for improving attack effectiveness. Random edge selection reduces ASR and provides no significant benefit in maintaining clean data accuracy.

\subsubsection{Impact of Poison Rate on Attack Effectiveness}
\begin{figure}[htbp]
    \centering
    \begin{subfigure}{0.48\linewidth}
        \centering
        \includegraphics[width=\linewidth]{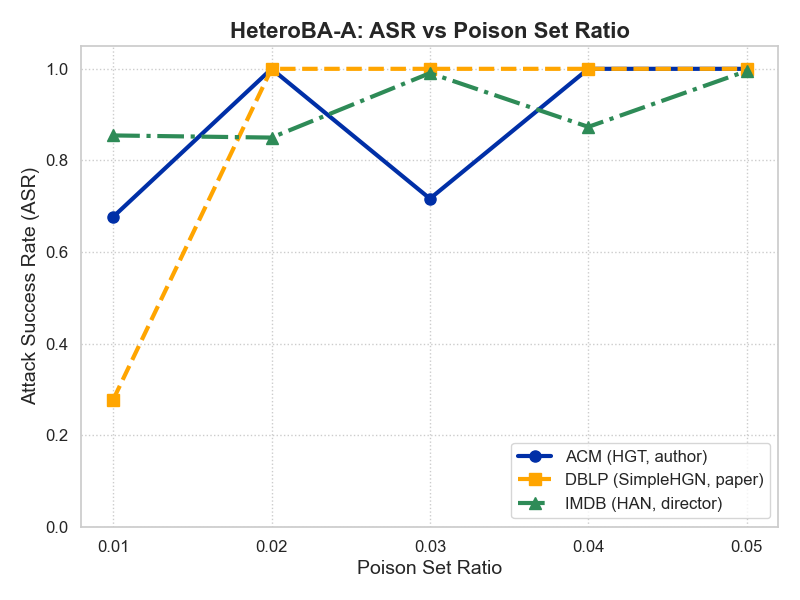}
        \caption{HeteroBA-A}
        \label{fig:heteroba_a}
    \end{subfigure}
    \hfill
    \begin{subfigure}{0.48\linewidth}
        \centering
        \includegraphics[width=\linewidth]{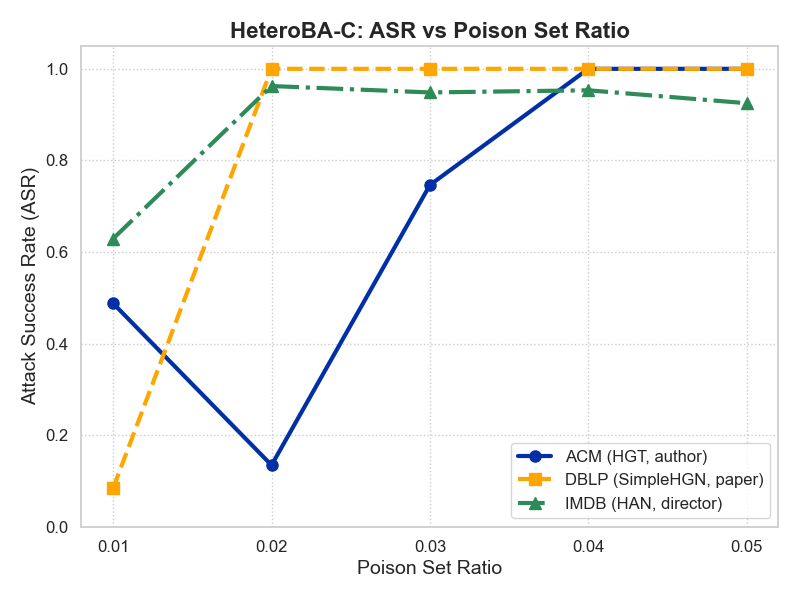}
        \caption{HeteroBA-C}
        \label{fig:heteroba_c}
    \end{subfigure}
    \caption{Comparison of attack success rates for HeteroBA-A and HeteroBA-C under different poison rates.}
    \label{fig:two_images}
\end{figure}
\vspace{-1em}
In this experiment, we explore how the poison rate influences attack success rate (ASR) across three representative scenarios to answer the \textbf{RQ4}, as depicted in the left figure (HeteroBA-A) and the right figure (HeteroBA-C) of Fig. 3. When the poison rate is only 0.01, the ASR is relatively low, but it increases markedly as the proportion of poisoned samples in the training set grows. This trend mirrors the phenomenon reported in \cite{system_crash}, where a complex system undergoes a sudden collapse at a specific critical point. Once the poison rate exceeds that threshold, the backdoor attack success rate tends to spike sharply, indicating heightened model vulnerability to the backdoor trigger within that range.

Notably, in some cases, even a very small poison rate (e.g., 0.01) can still yield a considerable level of attack effectiveness. This suggests that when the backdoor trigger is well-aligned with the model architecture or the underlying data distribution, only a modest number of poisoned samples is required for the model to learn the backdoor features and achieve a high ASR. These findings underscore the insidious nature of backdoor attacks and highlight the pressing need for robust defensive strategies.

\subsubsection{Data visualization}
In the Fig.~\ref{fig:2x3grid},
the “Attention Distribution” panels show that the victim model’s attention scores are heavily concentrated on a few influential auxiliary-type nodes within the second-hop neighborhood, following a long-tailed pattern. These nodes capture most of the attention weight, while the majority of nodes contribute minimally. In the “Embedding Cluster” figures, highlighted nodes—top-ranked by average cosine similarity—are positioned near dense cluster centers, reflecting strong semantic or structural similarity with neighboring nodes.

These patterns reveal that a small set of auxiliary-type nodes plays a key role in propagating adversarial triggers. Their high attention scores and central cluster positions make them effective conduits for influencing target node representations. This enables more successful backdoor attacks, as triggers align with the graph’s inherent structure. These findings address \textbf{RQ5} by showing that these key nodes can efficiently transmit adversarial signals to target nodes.

\section{Conclusion and future work }
In this paper, we introduce HeteroBA, the first backdoor attack framework tailored for HGNNs in node classification. By injecting trigger nodes and forming targeted connections with both primary and auxiliary neighbors through two distinct strategies, HeteroBA misleads the model into predicting a designated target class while maintaining clean data performance. Extensive experiments on benchmark datasets and various HGNN architectures reveal that HeteroBA achieves high attack success rates with strong stealthiness.

Future work will explore extending the attack to other tasks, such as recommendation systems and graph classification, and developing effective defense strategies against backdoor attacks on heterogeneous graphs.

\bibliographystyle{ACM-Reference-Format}
\bibliography{sample-base}

\appendix

\begin{table*}[ht!]
\centering
\scriptsize
\caption{Remaining attack results.}
\label{tab:appendix-triggers}
\begin{tabular}{c|c|c|c|ccccc|ccccc}
\hline
\multirow{2}{*}{\textbf{Dataset}} 
 & \multirow{2}{*}{\textbf{Victim Model}} 
 & \multirow{2}{*}{\textbf{Class}} 
 & \multirow{2}{*}{\textbf{Trigger}} 
 & \multicolumn{5}{c|}{\textbf{ASR}} 
 & \multicolumn{5}{c}{\textbf{CAD}} \\
\cline{5-14}
 &  &  & 
 & HeteroBA-C & HeteroBA-A & HeteroBA-R & CGBA & UGBA
 & HeteroBA-C & HeteroBA-A & HeteroBA-R & CGBA & UGBA \\
\hline
\multirow{9}{*}{\textbf{ACM}}
 & \multirow{3}{*}{HAN}
   & 0 & \multirow{3}{*}{field}
     & 0.3167 & 0.7612 & 0.9021 & \textbf{0.9420} & 0.0697
     & \textbf{-0.0072} & 0.0508 & $-0.0419^\dagger$ & 0.0160 & 0.0122 \\
 &  & 1 & 
     & 0.6667 & \textbf{0.6716} & $0.7960^\dagger$ & 0.5970 & 0.1178
     & \textbf{0.0066} & 0.0513 & $-0.0171^\dagger$ & 0.1109 & 0.0187 \\
 &  & 2 &
     & 0.0829 & \textbf{1.0000} & 0.9967 & 0.8126 & 0.0398
     & 0.0006 & 0.1424 & -0.0215 & \textbf{-0.0375} & 0.0000 \\
\cline{2-14}
 & \multirow{3}{*}{HGT}
   & 0 & \multirow{3}{*}{field}
     & 0.7164 & 0.4859 & 0.6020 & 0.9436 & \textbf{0.9867}
     & -0.0105 & -\textbf{0.0105} & 0.0016 & -0.0022 & -0.0038 \\
 &  & 1 & 
     & 0.7861 & 0.8060 & 0.6733 & 0.8905 & \textbf{0.9851}
     & 0.0055 & -0.0011 & -0.0028 & 0.0122 & \textbf{-0.0061} \\
 &  & 2 &
     & 0.6020 & 0.7131 & 0.8275 & 0.9005 & \textbf{0.9469}
     & 0.0039 & 0.0033 & 0.0055 & 0.0083 & \textbf{0.0006} \\
\cline{2-14}
 & \multirow{3}{*}{SimpleHGN}
   & 0 & \multirow{3}{*}{field}
     & 0.9718 & \textbf{1.0000} & 0.9983 & 0.9602 & \textbf{1.0000}
     & 0.0094 & 0.0160 & -0.0022 & -0.0022 & \textbf{-0.0660} \\
 &  & 1 &
     & 0.9735 & \textbf{1.0000} & 0.9884 & 0.9303 & \textbf{1.0000}
     & -0.0061 & -0.0022 & 0.0033 & 0.0000 & \textbf{-0.0182} \\
 &  & 2 &
     & 0.9685 & \textbf{1.0000} & 1.0000 & 0.9038 & \textbf{1.0000}
     & 0.0033 & 0.0066 & 0.0066 & 0.0033 & \textbf{-0.0701} \\
\hline
\multirow{9}{*}{\textbf{IMDB}}
 & \multirow{3}{*}{HAN}
   & 0 & \multirow{3}{*}{actor}
     & 0.2866 & \textbf{0.6854} & 0.5358 & 0.5618 & 0.2087
     & 0.0208 & \textbf{0.0016} & 0.0099 & 0.0037 & 0.0364 \\
 &  & 1 &
     & \textbf{0.9751} & 0.7212 & 0.5545 & 0.4523 & 0.2991
     & \textbf{-0.0187} & \textbf{-0.0187} & -0.0041 & -0.0119 & 0.0047 \\
 &  & 2 &
     & 0.8178 & \textbf{0.8910} & 0.5997 & 0.4992 & 0.3832
     & -0.0031 & \textbf{-0.0073} & 0.0088 & 0.0010 & 0.0114 \\
\cline{2-14}
 & \multirow{3}{*}{HGT}
   & 0 & \multirow{3}{*}{actor}
     & 0.6511 & \textbf{0.8006} & 0.6230 & 0.4851 & 0.5109
     & 0.0140 & 0.0026 & 0.0052 & \textbf{-0.0104} & 0.0291 \\
 &  & 1 &
     & 0.8022 & \textbf{0.8240} & 0.7227 & 0.4147 & 0.7757
     & 0.0145 & 0.0244 & $-0.0078^\dagger$ & 0.0130 & \textbf{0.0026} \\
 &  & 2 &
     & 0.7290 & \textbf{0.8286} & 0.6075 & 0.4523 & 0.6807
     & 0.0109 & 0.0145 & 0.0036 & \textbf{-0.0015} & 0.0182 \\
\cline{2-14}
 & \multirow{3}{*}{SimpleHGN}
   & 0 & \multirow{3}{*}{actor}
     & 0.8660 & \textbf{0.9829} & 0.7492 & 0.3881 & 0.8427
     & -0.0005 & -0.0047 & 0.0052 & \textbf{-0.0244} & -0.0016 \\
 &  & 1 &
     & \textbf{0.9346} & 0.8614 & $0.9455^\dagger$ & 0.3850 & 0.9611
     & 0.0015 & -0.0062 & 0.0146 & \textbf{-0.0130} & 0.0322 \\
 &  & 2 &
     & 0.8988 & \textbf{0.9408} & 0.8084 & 0.3474 & 0.9330
     & 0.0047 & \textbf{-0.0057} & -0.0031 & 0.0156 & 0.0047 \\
\hline
\end{tabular}
\end{table*}

\section{Time complexity analysis}
\label{appendix_sec:time complexity analysis}
The time complexity of the algorithm is primarily determined by the graph size, the number of target nodes for trigger insertion, and the selection strategy of auxiliary nodes. Let \( n = |\mathcal{V}| \) denote the total number of nodes in the graph, and \( m = |\mathcal{E}| \) the total number of edges. Define \( \mathcal{V}_{t_p} \) as the set of nodes of type \( t_p \), with a size of \( n_{t_p} \). The set of target nodes for trigger insertion is denoted as \( \mathcal{V}^{(p)} \subseteq \mathcal{V}_{t_p} \), with size \( p = |\mathcal{V}^{(p)}| \). Additionally, let \( \mathcal{V}_{\mathrm{aux}} = \bigcup_{t_b \in \mathcal{T}_{\mathrm{aux}}} \mathcal{V}_{t_b} \) be the set of auxiliary type nodes, with size \( n_{\mathrm{aux}} \).

The algorithm first filters out nodes of the target type that do not match a specific label and identifies relevant nodes in their neighborhood. A single-hop neighbor search typically has a complexity of at most \( O(m) \), depending on the adjacency structure. If multi-hop neighborhoods are considered, this step can be viewed as a breadth-first search (BFS) on the relevant subgraph, with an upper bound of \( O(n + m) \). Following this, the average degree information for different types of nodes is computed, which involves scanning specific sets of nodes or edges. Without index optimization, this step also has a complexity of \( O(n + m) \).

During the trigger insertion process, each target node is assigned a new trigger node, and its feature values are sampled accordingly. If the features are discrete, a simple Bernoulli sampling method is applied, which has a complexity of \( O(1) \). For continuous features using kernel density estimation (KDE), the complexity depends on whether the KDE model is pre-trained. If a pre-trained model is available, the complexity per sample is \( O(1) \) or \( O(\log n) \). The newly inserted trigger node must be connected to the target node and auxiliary nodes. To ensure proper connectivity, auxiliary nodes are selected using an attention-based or clustering-based sorting mechanism. If sorting is applied, the complexity can be as high as \( O(n_{\mathrm{aux}} \log n_{\mathrm{aux}}) \). If clustering is used, the complexity depends on the specific clustering algorithm, but in most cases, it remains close to \( O(n_{\mathrm{aux}} \log n_{\mathrm{aux}}) \). Since this process is executed for each target node, the cumulative complexity becomes \( O(p \cdot n_{\mathrm{aux}} \log n_{\mathrm{aux}}) \). After selecting the auxiliary node set, adding the corresponding edges incurs a linear complexity with respect to the number of selected nodes. If the maximum number of connections per node is bounded by \( D \), this step has a complexity of \( O(D) \).
\begin{algorithm}[t]
\small
\caption{HeteroBA Overall Algorithm}
\label{alg:HeteroBA}
\begin{algorithmic}[1]
\STATE \textbf{Input:} $G=(\mathcal{V},\mathcal{E},X)$, $t_p,t_{tr},\mathcal{T}_{\mathrm{aux}}$, $y_t$, $\mathcal{V}^{(p)}\subseteq \mathcal{V}_{t_p}$
\STATE \textbf{Output:} $\widetilde{G}=(\widetilde{\mathcal{V}},\widetilde{\mathcal{E}},\widetilde{X})$

\STATE Identify $\mathcal{V}_{\neg y_t} \leftarrow \{v \in \mathcal{V}_{t_p}\mid y_v \neq y_t\}$
\STATE $\mathcal{V}'_{t_{tr}} \leftarrow \mathrm{Neigh}(\mathcal{V}_{\neg y_t},\,r_{t_p,t_{tr}})$
\STATE Compute average degrees $\{d_{t_b}\}$ for edges $(t_{tr},t_b)$, $\forall t_b \in \mathcal{T}_{\mathrm{aux}}$
\STATE $\mathcal{E}^{(\mathrm{new})} \leftarrow \varnothing,\quad X^{(\mathrm{new})}\leftarrow \varnothing$

\FOR{each $v \in \mathcal{V}^{(p)}$}
    \STATE Insert a new trigger node $u$
    \STATE \textbf{if} feature is continuous: sample $x_u^{(\mathrm{new})}$ via KDE
    \STATE \textbf{else} (binary): sample $x_u^{(\mathrm{new})}$ via Bernoulli
    \STATE $\mathcal{E}^{(\mathrm{new})} \leftarrow \mathcal{E}^{(\mathrm{new})}\cup\{(u,v),(v,u)\}$
    \STATE Let $W \leftarrow \{\text{top-}d_{t_b}\text{ in each }\mathcal{V}_{\mathrm{aux}}\;\text{(via attention/clustering)}\}$
    \STATE $\mathcal{E}^{(\mathrm{new})} \leftarrow \mathcal{E}^{(\mathrm{new})}\cup\{(u,w),(w,u)\,\mid\,w \in W\}$
    \STATE $X^{(\mathrm{new})} \leftarrow X^{(\mathrm{new})}\cup\{\,x_u^{(\mathrm{new})}\}$
\ENDFOR

\STATE $\widetilde{\mathcal{V}} \leftarrow \mathcal{V}\cup\{\text{all new }u\}$
\STATE $\widetilde{\mathcal{E}} \leftarrow \mathcal{E}\cup\mathcal{E}^{(\mathrm{new})}$
\STATE $\widetilde{X} \leftarrow
\begin{bmatrix}
X\\[2pt]
X^{(\mathrm{new})}
\end{bmatrix}
$

\RETURN $\widetilde{G}=(\widetilde{\mathcal{V}},\widetilde{\mathcal{E}},\widetilde{X})$
\end{algorithmic}
\end{algorithm}

Once all trigger nodes have been constructed, they must be merged into the original graph, including updating the node set, edge set, and feature matrix. The number of newly added nodes is \( p \), and the number of new edges is typically at most \( O(pD) \), leading to a complexity of \( O(p) \) or \( O(pD) \). Overall, the total time complexity of the algorithm is given by:
\[
O(n + m) + O\bigl(p \cdot n_{\mathrm{aux}} \log n_{\mathrm{aux}}\bigr).
\]

\begin{table}[t]
\centering
\small
\caption{Training Parameters and Model-Specific Hyperparameters}
\label{tab:combined_params_simplified}
\begin{tabular}{ll}
\toprule
\multicolumn{2}{c}{\textbf{Key Training Parameters}} \\
\midrule
\textbf{Parameter}      & \textbf{Value}      \\
\midrule
Loss function           & Cross Entropy       \\
Optimizer               & AdamW               \\
Epochs                  & 400                 \\
Learning rate           & 1e-3                \\
Scheduler               & OneCycleLR          \\
Dropout                 & 0.2                 \\
Weight decay            & 1e-4                \\
Gradient Clipping       & 1.0                 \\
\midrule
\multicolumn{2}{c}{\textbf{Model-Specific Hyperparameters}} \\
\midrule
\textbf{Model}  & \textbf{Hyperparameters (Value)} \\
\midrule
HGT           & Hidden Units: 64; Layers: 8; Heads: 4 \\
HAN           & Hidden Units: 64; Heads: 4 \\
SimpleHGN     & Hidden Units: 64; Heads: 8; Layers: 4; \\
\bottomrule
\end{tabular}
\end{table}

\section{Other training parameters}
\label{appendix_sec:other training parameters}
The training parameters are shown in Table \ref{tab:combined_params_simplified}.

\section{Other attack results}
The remaining attack results are presented in Table \ref{tab:appendix-triggers}.

\section{Pseudocode of HeteroBA}
\label{appendix_sec: pseudocode}
The pseudocode of HeteroBA is in Algorithm \ref{alg:HeteroBA}

\end{document}